# DrugRAG: Enhancing Pharmacy LLM Performance Through A Novel Retrieval-Augmented Generation Pipeline

Houman Kazemzadeh [1], Kiarash Mokhtari Dizaji [2], Seyed Reza Tavakoli [3], Farbod Davoodi [4], MohammadReza KarimiNejad [5], Parham Abed Azad [5], Fatemeh Latifi [6], Ali Sabzi [5], Armin Khosravi [5], Siavash Ahmadi [7], Babak Khalaj [8], Mohammad Hossein Rohban [5], Gholamali Aminian [9], Zohreh Amoozgar [10], Tahereh Javaheri [11, *]

1 Department of Medicinal Chemistry, Faculty of Pharmacy, Tehran University of Medical Sciences, Tehran, Iran
2 Department of Computer Sciences, Faculty of Mathematics and Computer Sciences, Amir Kabir University of Technology, Tehran, Iran
3 Department of Mathematical Sciences, Sharif University of Technology, Tehran, Iran
4 Department of Computer Sciences, Missouri University of Science and Technology, Rolla, MO, USA
5 Department of Computer Engineering, Sharif University of Technology, Tehran, Iran
6 Department of Faculty of Interdisciplinary Science and Technology, Tarbiat Modares University, Tehran, Iran
7 Electronics Research Institute, Sharif University of Technology, Tehran, Iran
8 Department of Electrical Engineering, Sharif University of Technology, Tehran, Iran
9 The Alan Turing Institute, London, United Kingdom
10 Department of Radiation Oncology, Massachusetts General Hospital & Harvard Medical School, Boston, USA
11 Health Informatics Lab, Metropolitan College, Boston University, Boston, USA
* Corresponding Author

## Abstract

In our study, we evaluated large language model (LLM) performance on pharmacy licensure-style question-answering tasks and developed an external knowledge integration method to improve accuracy. We benchmarked ten LLMs with varying parameter sizes (8 billion to 70+ billion) using a 141-question pharmacy dataset, measuring baseline accuracy without modification. Baseline performance ranged from 46% to 92%, with GPT-5 (92%) and o3 (89%) achieving the highest scores, while smaller open-source models showed substantially lower performance. We then developed DrugRAG, a three-step retrieval-augmented generation (RAG) pipeline that retrieves structured, evidence-based drug information and augments model prompts with contextual pharmacological evidence, operating externally and requiring no changes to model architecture or parameters. DrugRAG increased accuracy across all five evaluated models, with gains ranging from 7 to 21 percentage points (e.g., Gemma 3 27B: 61.0% to 71%, Llama 3.1 8B: 46% to 67%). McNemar analyses demonstrated statistically significant paired improvements primarily in smaller and mid-sized open-source models. These findings demonstrate that integrating structured external drug knowledge via DrugRAG can improve LLM performance on pharmacy-focused question-answering tasks without modifying the underlying models, providing a practical pipeline for enhancing evidence-based pharmacy-focused AI applications.

## 1. Introduction

Pharmacists must master accurate drug selection, dose calculations, and context-specific decision-making, presenting distinct challenges for AI systems in pharmacy education and practice [1]. While large language models have created new opportunities for AI-supported learning in pharmacy and healthcare [2,3], and general-purpose models such as GPT-4 show high performance on medical education exams [4,5], pharmacy-specific capabilities require rigorous evaluation before deploying LLMs as tutoring systems or study aids.

The NAPLEX (North American Pharmacist Licensure Examination), administered by the National Association of Boards of Pharmacy, is the standard licensure exam for pharmacy graduates. It evaluates entry-level competency across five domains: Foundational Knowledge for Pharmacy Practice, Medication Use Process, Person-Centered Assessment and Treatment Planning, Professional Practice, and Pharmacy Management and Leadership [6,7].

Recent benchmarks reveal significant performance gaps among language models on NAPLEX-style questions. Angel et al. tested GPT-4, Bard, and GPT-3 on 137 questions, with GPT-4 achieving 78.8% accuracy [8]. Ehlert et al. found GPT-4 reaching 87% on McGraw-Hill questions while smaller models trailed at 60-68% [9]. A follow-up study by Angel et al. compared six models on 225 items, showing GPT-4 at 87.1% but Llama-2-70B-chat reaching only 48% [10].

These findings demonstrate that model scale and targeted fine-tuning are critical for pharmacy-domain reasoning.

However, most prior studies focused primarily on benchmarking model accuracy and did not investigate whether external structured drug knowledge could systematically improve pharmacy-domain reasoning without retraining the underlying models.

We address two key questions: How do current LLMs of varying sizes perform on pharmacy questions, and can external knowledge integration improve their accuracy without retraining?

We benchmarked ten LLMs on a 141-question pharmacy dataset and developed DrugRAG, a retrieval-augmented generation pipeline that augments model prompts with structured evidence-based pharmacological information. Our approach works with models as-is, requiring no modification to their architecture or parameters.

This study demonstrates that external structured knowledge integration can improve LLM performance on pharmacy-specific question-answering tasks.

## 2. Methods

### 2.1. Study Design and Question Set

We evaluated ten language models using 141 multiple-choice questions obtained from PharmacyExam, a publicly accessible NAPLEX preparation resource [11]. These questions consist of clinical vignettes with four or five answer options and cover a range of pharmacotherapy and pharmacy practice topics, including dosing, contraindications, drug–drug interactions, and therapeutic decision-making.

The dataset has several limitations. It originates from a single commercial preparation source and lacks formal psychometric properties such as item difficulty or discrimination indices. Additionally, while the questions span multiple areas of pharmacy practice, we did not formally validate their mapping to the five NAPLEX competency domains [7]. Therefore, this dataset represents a practical but not exhaustive sample of pharmacy question-answering tasks. Despite these constraints, it provides a clinically relevant benchmark for evaluating model performance on exam-style reasoning.

We prompted each model with the case scenario and options, mirroring how human test-takers engage with the exam. We scored only the final answer choice (A, B, C, or D), calculating accuracy as the percentage of correct responses out of 141 questions. An answer was correct if it matched the official key. Ambiguous answers or choices outside the provided options were marked incorrect.

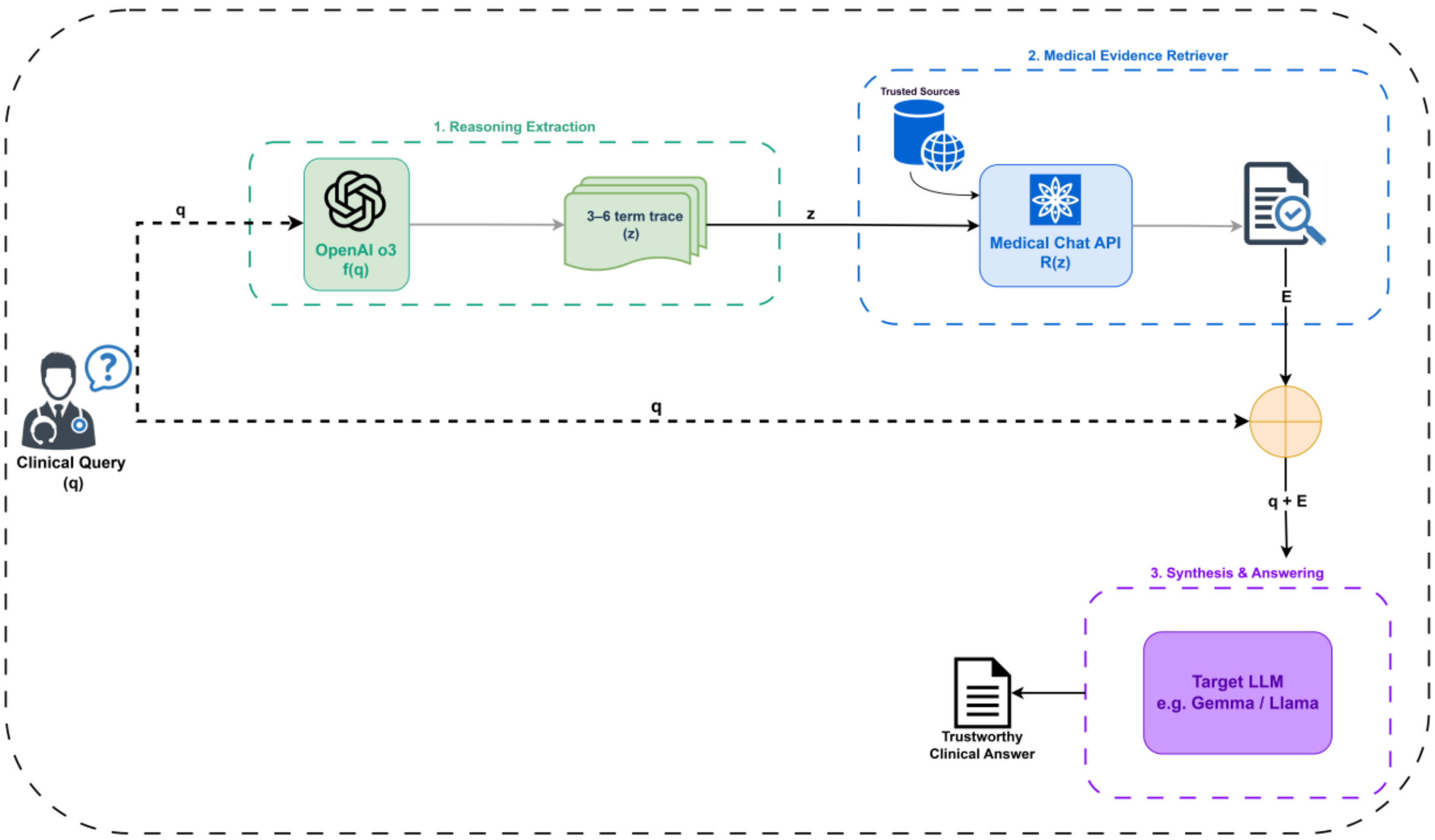

**Figure 1. DrugRAG architecture for medication question answering.**
A clinical query (q) is first passed to o3, which extracts a focused 3–6 term reasoning trace. These terms guide a medical evidence retriever that queries trusted drug information sources through the Medical Chat API, producing a structured evidence snippet (E). The target LLM receives the augmented input (q+E) and synthesizes the final clinical answer.

### 2.2. Evaluated Language Models

We selected ten existing models for evaluation based on public availability, biomedical relevance, and parameter scale diversity. Model parameters (measured in billions, denoted as "B") represent the number of trainable weights in the neural network; larger parameter counts generally indicate greater model capacity and knowledge. Our selection ranged from 8B (8 billion parameters) to over 70B parameters, as well as proprietary models with undisclosed sizes. We tested all models in their original form without any modifications to architecture, parameters, or fine-tuning.

Open-source models (Bio-Medical Llama 3 8B [12], Llama 3.1 8B [13], Gemma 3 27B [14]) provided accessible baselines for biomedical research. Proprietary models (GPT-4o [15], GPT-5 [16], o3 [17], o4 Mini [18], Gemini 2.0 Flash [19], Gemini 3 Pro [20], Claude Opus 4.5 [21]) represented top-tier performance. Proprietary models were identified using the model names or API endpoint labels available at the time of evaluation. In some cases, these names reflect provider-side naming conventions or preview releases (e.g., internal or API-facing identifiers) that may differ from later standardized public documentation. Therefore, access dates are reported to contextualize model behavior and ensure reproducibility within the constraints of evolving LLM APIs.

We ensured a fair comparison by using identical prompts across all models. For open-source models, we standardized generation settings (temperature = 0.2, top_p = 1.0, frequency penalty = 0.0, presence penalty = 0.0, max_tokens = 512) to minimize stochastic variation and isolate model reasoning performance. For proprietary models (including GPT-4o, GPT-5, o3, o4 Mini, Gemini models, and Claude Opus 4.5), we used the default API generation parameters provided by each vendor, as these settings are not fully user-controllable or consistently exposed across platforms. This approach reflects the practical constraints of proprietary API-based systems and ensures reproducibility within the constraints of each system.

***Table 1. Baseline Accuracy of Large Language Models***

| Model | Parameter Size | Accuracy (% Correct) | 95% CI (%) |
|---|---|---|---|
| **Bio-Medical Llama 3** | 8B (8 billion) | 46 | 38–54 |
| **Llama 3.1** | 8B (8 billion) | 46 | 38–54 |
| **Gemma 3** | 27B (27 billion) | 61 | 53–69 |
| **Gemini 2.0 (Flash)** | Not disclosed | 72 | 64–79 |
| **Gemini 3 (Pro)** | Not disclosed | 75 | 67–82 |
| **o4 Mini** | Not disclosed | 76 | 68–82 |
| **GPT-4o** | Not disclosed | 81 | 74–86 |
| **Claude Opus 4.5** | Not disclosed | 87 | 81–92 |
| **o3** | Not disclosed | 89 | 82–93 |

| GPT-5 | Not disclosed | 92 | 87–96 |
|---|---|---|---|

Baseline accuracy represents performance without any external knowledge augmentation. Parameter sizes are approximate; "Not disclosed" indicates proprietary specifications. B = billion parameters.

### 2.3. DrugRAG Pipeline Development

We developed DrugRAG, a three-stage retrieval-augmented generation pipeline that integrates structured drug knowledge with existing LLMs (Figure 1). This pipeline operates externally and requires no modification of model architecture, parameters, or training.

**Step 1 – Reasoning Extraction:** We used o3 to extract 3–6 key medical concepts from each clinical query. o3 was selected because it produced concise, stable reasoning traces suitable for evidence retrieval. The goal was not to evaluate o3 as an answer-generating model, but to generate a reproducible and semantically focused retrieval query. We acknowledge that using a high-performing reasoning model may improve the conceptual framing of each question; therefore, observed DrugRAG gains reflect the full three-step pipeline rather than evidence retrieval alone.

**Step 2 – Evidence Retrieval:** We queried the Medical Chat API with the reasoning trace [22], retrieving structured drug information (maximum 200 words) from trusted pharmacological references and drug databases. The API returns evidence in a structured format including drug name, indication, dosage limits, contraindications, and other relevant details.

**Step 3 – Augmented Prompting:** We combined the original question with the structured evidence block and prompted the target LLM. Instructions explicitly directed the model to ground its answer in the provided evidence, reducing unsupported responses and aligning answers with medical consensus.

This approach enables external evidence augmentation for pharmacy question-answering without retraining or modifying the underlying model.

#### 2.3.1. Evidence Source Selection

We selected Medical Chat as the evidence retrieval layer after evaluating several open and semi-open drug information resources, including OpenFDA [23], DrugCentral [24], DrugBank [25], RxNorm [26], and other publicly accessible drug references. Medical Chat functioned as an external, LLM-assisted retrieval interface that aggregates information from trusted pharmacological sources and returns concise, clinically relevant evidence. This distinction is important: DrugRAG does not compare Medical Chat as a competing model, but examines whether structured external drug evidence can improve the performance of independent target LLMs.

The external evidence was organized around clinically actionable pharmacotherapy elements: drug name, indication, dosage limits, contraindications, adverse effects, drug–drug interactions, and context-specific details. This structured format makes retrieved evidence easier for target LLMs to interpret and apply.

We considered directly querying open-source databases without an intermediate retrieval model, but found them insufficient for broad pharmacy question-answering. OpenFDA provides drug labels and adverse event reports, but outputs require substantial processing before conversion to clinical guidance. DrugCentral contains detailed pharmacological information but is often chemistry-oriented or incomplete for scenario-based clinical decisions. DrugBank is comprehensive but requires a license and is optimized for research rather than clinical question synthesis. RxNorm is useful for drug naming and synonym mapping but does not provide clinical recommendations or therapeutic guidance. Resources like Drugs.com and the Merck Manual offer drug monographs but are not designed as structured APIs for real-time, question-specific retrieval.

We also considered professional systems like UpToDate and Lexidrug/Lexicomp, which would be highly relevant for pharmacotherapy reasoning, but their proprietary nature and restricted API access limited feasibility. Given these constraints, Medical Chat served as a practical LLM-assisted evidence retrieval interface that synthesizes relevant information from trusted sources into concise evidence blocks. DrugRAG should therefore be interpreted as an external evidence-augmentation pipeline using LLM-assisted retrieval, rather than a system based solely on a neutral structured database. This framing reflects practical constraints of accessing high-quality clinical drug knowledge while preserving the study's central aim: evaluating whether structured pharmacotherapy evidence improves LLM performance on pharmacy question-answering tasks.

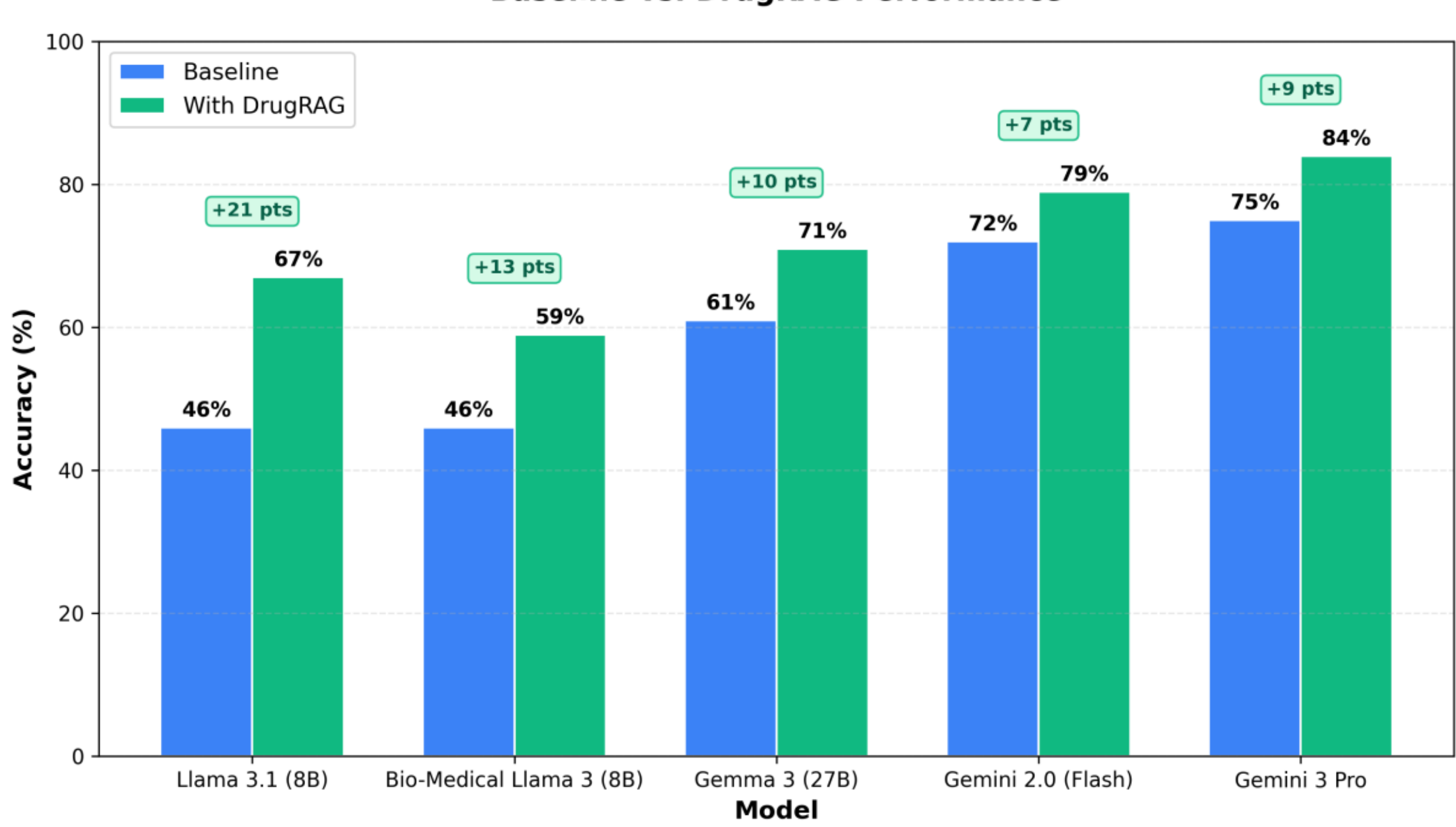

**Figure 2. Effect of DrugRAG on LLM accuracy in pharmacy question answering.**
DrugRAG produced positive accuracy changes across all evaluated models. Larger relative improvements were observed in smaller open-source models. Statistically significant paired improvements were primarily observed in smaller and mid-sized open-source models.

### 2.4. Statistical Analysis

All analyses were performed using Python 3.11 with NumPy, Pandas, Matplotlib, Seaborn, and Statsmodels. Model performance was evaluated using accuracy, defined as the proportion of correctly answered questions out of the total set (n = 141).

Because all models were evaluated on the same set of questions, the resulting outcomes are paired rather than independent. Therefore, we employed McNemar’s test to assess the statistical significance of differences between baseline and DrugRAG conditions for each model. McNemar’s test evaluates paired binary outcomes and is appropriate for determining whether the intervention (DrugRAG) significantly changed model responses on the same questions.

For each model, we computed the number of discordant pairs, defined as cases where DrugRAG corrected an initially incorrect response (b) and cases where DrugRAG changed a correct response to an incorrect one (c). Statistical significance was assessed using McNemar’s test with continuity correction, and $p$-values $< 0.05$ were considered statistically significant.

To quantify uncertainty in accuracy estimates, we calculated 95% confidence intervals using the Wilson score method for binomial proportions. Confidence intervals were reported for all baseline and DrugRAG accuracy values presented in Tables 1 and 2. All statistical tests were two-sided.

## 3. Results and Discussion

### 3.1. Baseline Model Performance Across Parameter Scales

Table 1 shows substantial variation in accuracy across models of different parameter sizes. These baseline results reflect each model's inherent capabilities without any external knowledge augmentation:

Bio-Medical Llama 3 8B and Llama 3.1 8B both achieved 46%, the lowest accuracy in our benchmark. This demonstrates the difficulty of pharmacy content for models at this scale (8 billion parameters), even with domain-specific pre-training.

Gemma 3 27B (27 billion parameters) reached 61%, a notable improvement over the 8B models. This suggests both scale and architecture contribute to pharmacy reasoning capability.

Gemini 2.0 Flash achieved 72%, showing that strong general-purpose reasoning can reach competitive performance on pharmacy questions.

Gemini 3 Pro reached 75%, indicating solid proficiency in medication-related decision-making and question interpretation.

o4 Mini scored 76%. Despite its smaller footprint, advanced fine-tuning appears to enhance test-taking strategies and logical reasoning.

GPT-4o obtained 81%, reflecting sophisticated language understanding and broad training data.

Claude Opus 4.5 scored 87%, demonstrating strong general-purpose reasoning and reliable performance on pharmacy-focused questions.

o3 scored 89%, demonstrating strong pharmacy-domain reasoning performance and showing how advanced general models can excel on pharmacy tasks.

GPT-5 achieved 92%, the highest performance in our evaluation. It demonstrates enhanced reasoning depth and represents the highest performance observed in our benchmark for pharmacy question-answering.

To better understand these baseline differences, we qualitatively examined common failure patterns across model outputs. Smaller models frequently exhibited formula-related errors, including forgetting standard pharmacotherapy equations, misapplying formulas, or producing different results across repeated attempts (e.g., in eAG calculation or weight-based dosing). We also observed contextual misunderstandings, in which a model recognized familiar medical terms but misinterpreted the task, for example, returning the therapeutic target for HbA1c rather than computing an eAG value. Another common issue involved logical and numerical inconsistencies: models made arithmetic mistakes, substituted incorrect parameters, mismatched units, or produced explanations that contradicted their final chosen answer, reflecting unstable multi-step reasoning. Finally, several models demonstrated medication-identification errors, such as repeatedly confusing Tavist (clemastine) with Tavist ND (loratadine).

Our findings also reveal that larger parameter-scale models generally demonstrate higher baseline performance when used without external knowledge augmentation. Among open-source models, only Gemma 3 27B exceeded 60% accuracy, while all models with fewer than 10 billion parameters remained below 50%. Bio-Medical Llama 3 and Llama 3.1 achieved similar performance despite different architectures, suggesting that domain-specific pretraining alone may not overcome size limitations at the 8B scale. We acknowledge that confounding factors in model design, architecture, and tuning prevent us from isolating individual contributions to baseline performance.

### 3.2. Performance Improvements Through DrugRAG

We applied our three-step RAG pipeline to five models with baseline accuracy below 75%. This threshold was selected as a pragmatic, competency-based cutoff rather than a direct NAPLEX equivalent. Prior work found a mean NAPLEX score of 104.7 among pharmacy graduates [27], with NABP identifying a scaled score of 75 or higher as passing. Since NABP states that scaled scores are not equivalent to percentage correct, we used 75% accuracy as a conservative heuristic to identify models with sub-threshold baseline performance for targeted DrugRAG evaluation.

#### 3.2.1. Individual Model Results

All improvements resulted entirely from integrating external knowledge, not from modifying the models themselves. We observed gains ranging from 7 to 21 percentage points, with smaller models showing the largest improvements (Figure 2).

**Llama 3.1 8B** improved from 46% to 67% (21-point gain). This substantial improvement suggests that 8-billion-parameter models can approach the baseline performance of much larger models when provided with structured drug knowledge through our external pipeline.

**Bio-Medical Llama 3 8B** increased from 46% to 59% (13-point improvement). While more modest than Llama 3.1, this still represents a substantial enhancement through external knowledge integration.

**Gemma 3 27B** rose from 61% to 71% (10-point gain). Even models with 27 billion parameters benefit meaningfully from externally provided structured evidence.

**Gemini 2.0 Flash** improved from 72% to 79% (7-point increase). High-performing models still show measurable gains from evidence augmentation.

**Gemini 3 Pro** increased from 75% to 84% (9-point gain). This suggests that incorporating evidence on targeted pharmacotherapy improves baseline performance even in strong models.

#### 3.2.2. Statistical Analysis and Interpretation

McNemar analyses demonstrated statistically significant paired improvements for Llama 3.1 8B, Bio-Medical Llama 3 8B, and Gemma 3 27B, while Gemini 2.0 Flash and Gemini 3 Pro showed positive but non-significant trends.

These findings suggest that external evidence augmentation may help compensate for missing or inconsistently recalled pharmacological information across multiple model scales without requiring model modification. Smaller models may benefit more from explicit pharmacological evidence because they exhibit lower baseline recall of domain-specific information. Larger models already possess broader knowledge but still benefit from grounding in authoritative, context-specific evidence delivered through our pipeline.

*Table 2. Performance with DrugRAG*

| Model | Baseline Accuracy (%) | 95% CI (%) | Accuracy with RAG (%) | 95% CI (%) | Improvement | b | c | p-value |
|---|---|---|---|---|---|---|---|---|
| Llama 3.1 (8B) | 46 | 38–54 | 67 | 59–74 | +21 points | 42 | 10 | <0.001 |
| Bio-Medical Llama 3 (8B) | 46 | 38–54 | 59 | 51–67 | +13 points | 30 | 10 | 0.003 |
| Gemma 3 (27B) | 61 | 53–69 | 71 | 63–78 | +10 points | 22 | 9 | 0.031 |
| Gemini 2.0 (Flash) | 72 | 64–79 | 79 | 71–85 | +7 points | 35 | 25 | 0.245 |
| Gemini 3 (Pro) | 75 | 67–82 | 84 | 77–89 | +9 points | 28 | 15 | 0.067 |

#### 3.2.3. Broader Applications and Advantages

DrugRAG offers advantages beyond this specific benchmark. Because it operates externally to the models, it can be integrated with diverse LLM architectures without requiring access to model weights, retraining, or architectural modifications.

The pipeline can be adapted for multiple applications:

- Drug-interaction checks
- Guideline-based question-answering
- Patient-specific dosing support
- Any application requiring concise, trusted evidence to inform model reasoning

#### 3.2.4. Design Benefits

The three-step design ensures:

- Evidence relevance through reasoning extraction
- Reliability through structured pharmacological evidence from trusted references
- Effective integration through structured prompting

All improvements are achieved through external evidence augmentation and structured prompting rather than retraining or modifying the underlying models.

## 4. Limitations

Our study has several important limitations that should be considered when interpreting results.

**Dataset and Question Format:** The 141-question benchmark covers broad pharmacy content, but we did not perform formal mapping to NAPLEX domains or item-difficulty analysis. This limits claims of comprehensive topical representation. Future work should incorporate validated multi-source question sets with established psychometric properties.

We evaluated only multiple-choice questions, which do not capture real-world complexity such as multi-step therapeutic planning, open-ended decision making, patient-specific medication optimization, or longitudinal clinical reasoning. Generalizability to such tasks remains untested.

**Model Selection and Evaluation Scope:** DrugRAG was applied only to models below our predefined operational threshold, not to all benchmarked models. Therefore, conclusions about its effectiveness in higher-performing proprietary models should be interpreted with caution.

**Evidence Retrieval Design:** Medical Chat was used as an LLM-assisted evidence-retrieval interface rather than a neutral, structured database. Although this reflected practical constraints in accessing professional clinical knowledge systems and allowed concise evidence synthesis, it may limit interpretability of which pipeline component contributed most to performance gains.

**Technical and Practical Constraints:** We accessed proprietary models via closed APIs without insight into internal architecture, training data, update schedules, or provider-side changes. Our RAG pipeline introduces practical constraints: API-based calls incur latency and cost, and the three-step setup may be difficult to scale in low-resource or real-time applications.

While prompts were standardized, we did not test robustness to paraphrased or reworded question variants, which could affect consistency in real-world use.

**Reasoning Extraction Confounding:** We used o3 for reasoning-trace extraction because it provided strong and reproducible concept extraction at the time of evaluation. However, this design may introduce confounding, as o3's conceptual framing could improve downstream retrieval quality independently of the evidence source itself. Future work should evaluate alternative reasoning extractors, including smaller open-source models and rule-based keyword extraction, to quantify the independent contribution of the reasoning-extraction stage.

## 5. Conclusion

**Key Findings:** We developed DrugRAG, a three-step retrieval-augmented generation pipeline that integrates structured pharmacological evidence externally. This approach yielded accuracy gains ranging from 7 to 21 percentage points across all evaluated models, with statistically significant paired improvements observed primarily in smaller and mid-sized open-source systems.

**Clinical and Technical Implications:** Our findings demonstrate that external knowledge integration can improve LLM performance on pharmacy-focused question-answering tasks without modifying underlying model architectures or parameters. This pipeline provides a scalable approach for incorporating evidence-based pharmacological information into pharmacy-oriented AI applications.

**Future Applications:** The external design of DrugRAG makes it adaptable to diverse pharmacy applications while maintaining compatibility with existing LLM systems, offering a practical pathway for enhancing AI-supported pharmacy education and clinical decision support.

## 6. Declaration of Competing Interest

The authors declare no conflicts of interest related to this study.

## 7. Funding

This research received no specific grant from any funding agency in the public, commercial, or not-for-profit sectors.

## 8. Declaration of Generative AI and AI-assisted technologies

During manuscript preparation, the authors used ChatGPT (GPT-4o) to improve readability and language. The authors reviewed and edited all content and take full responsibility for the publication.